\documentclass[letterpaper, 10 pt, conference]{ieeeconf}
\IEEEoverridecommandlockouts                               
\usepackage{breqn}
\usepackage{cuted}
\usepackage{xcolor}
\usepackage{capt-of}
\usepackage{amsfonts}
\usepackage{hyperref}
\usepackage{multirow}
\usepackage{mathtools}
\usepackage{dblfloatfix}
\usepackage[font=small, labelfont=bf]{caption}
\usepackage[ruled, lined, linesnumbered, commentsnumbered, longend, noend]{algorithm2e}

\usepackage{fancyhdr} 
\newcommand{\minimize}{\mathop{\mathrm{min}}} 

\newcommand{\boldasterisk}[1]{\textbf{#1}$^{\ast}$} 
\newcommand{\bolddagger}[1]{\textbf{#1}$^{\dag}$} 
\newcommand{\boldssign}[1]{\textbf{#1}$^{\S}$} 

\definecolor{pink}{RGB}{255, 192, 203}

\title{\LARGE \bf V3D-SLAM: Robust RGB-D SLAM in Dynamic Environments with\\3D Semantic Geometry Voting}

\author{\large Tuan Dang \hspace{45pt} Khang Nguyen \hspace{45pt} Manfred Huber {\footnotesize \thanks{All authors are with the Learning and Adaptive Robotics Laboratory, Department of Computer Science and Engineering, University of Texas at Arlington, Arlington, TX 76013, USA. (emails: \href{mailto:tuan.dang@uta.edu}{\text{tuan.dang@uta.edu}}, \href{mailto:khang.nguyen8@mavs.uta.edu}{\text{khang.nguyen8@mavs.uta.edu}}, \href{mailto:huber@cse.uta.edu}{\text{huber@cse.uta.edu}})}}}

\begin{document}

\maketitle 

\thispagestyle{empty}
\pagestyle{empty}

\begin{strip}
    \vspace{-58pt}
    \centering
    \includegraphics[width=1.00\linewidth]{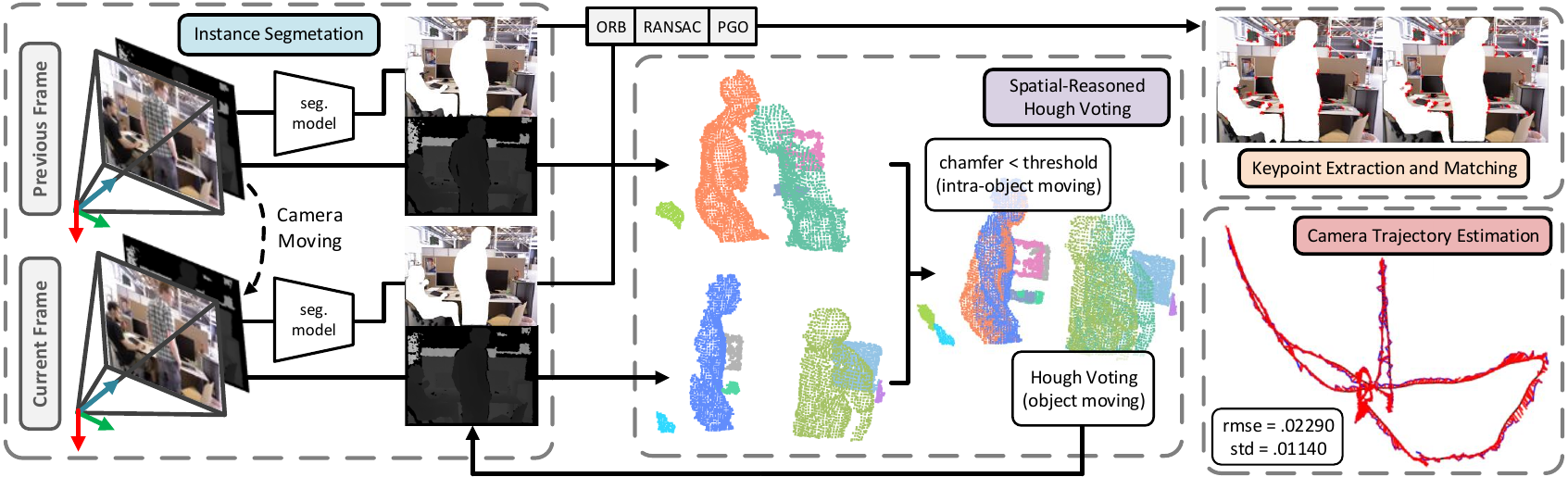}
    \vspace{-12pt}
    \captionof{figure}{Overview of \textit{V3D-SLAM}: improving the robustness of RGB-D SLAM in dynamic indoor environments, including instance segmentation coupled with RGB-based feature extraction (Sec. \ref{sec:segmentation_extraction}), sensor noises and segmentation outlier rejection (Sec. \ref{sec:outlier_rejection}), and spatial-reasoned Hough voting mechanism for dynamic 3D objects (Sec. \ref{sec:geometry_voting}), resulting in camera trajectory estimation (Sec. \ref{sec:evaluation}).}
    \vspace{-10pt}
    \label{fig:overview_method}
\end{strip}

\begin{abstract}
Simultaneous localization and mapping (SLAM) in highly dynamic environments is challenging due to the correlation complexity between moving objects and the camera pose. Many methods have been proposed to deal with this problem; however, the moving properties of dynamic objects with a moving camera remain unclear. Therefore, to improve SLAM's performance, minimizing disruptive events of moving objects with a physical understanding of 3D shapes and dynamics of objects is needed. In this paper, we propose a robust method, \textit{V3D-SLAM}, to remove moving objects via two lightweight re-evaluation stages, including identifying potentially moving and static objects using a spatial-reasoned Hough voting mechanism and refining static objects by detecting dynamic noise caused by intra-object motions using Chamfer distances as similarity measurements. Our experiment on the TUM RGB-D benchmark on dynamic sequences with ground-truth camera trajectories showed that our methods outperform the most recent state-of-the-art SLAM methods. Our source code is available at \href{https://github.com/tuantdang/v3d-slam}{https://github.com/tuantdang/v3d-slam}.
\end{abstract}

\section{Introduction}

Visual simultaneous localization and mapping (vSLAM), an important study of robotics, essentially relies on visual information from the camera to localize itself and build a map of the environment. With the emergence of low-cost cameras, vSLAM captures tremendous attention from the research community, especially RGB-D-based SLAM, which is one of the most popular choices for its availability and appropriateness for indoor scenes. A number of notable works have been proposed, such as Dense-SLAM \cite{kerl2013dense} and ORB-SLAM \cite{mur2015orb}; however, like traditional SLAM methods \cite{thrun2001robust, grisetti2007improved}, these approaches implicitly assume that the environment is static, where frame-to-frame extracted keypoints are matched, often failing to deal with dynamic scenes when objects and the camera are moving simultaneously.

\begin{figure*}[t]
    \centering
    \includegraphics[width=1.00\linewidth]{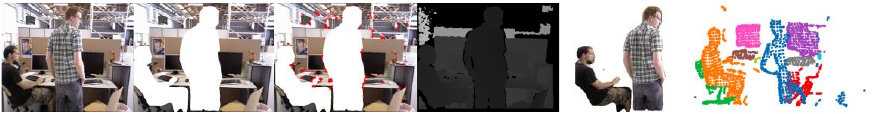}
    \vspace{-14pt}
    \caption{Segmentation of hypothesized moving objects with key points on static objects and background, and point clouds of object instances.}
    \vspace{-17pt}
    \label{fig:seg_feature}
\end{figure*}

Although vSLAM has been extensively studied with help from recent advancements in computer vision, especially with deep neural networks, some issues have not been well-addressed due to the dynamics of the environment. DS-LAM \cite{yu2018ds} uses a segmentation model and optical flow to detect moving objects and consider these moving objects as noises to be removed. TRS-SLAM \cite{ji2021towards} deals with objects on the training dataset while segmenting the depth image via k-means clustering to handle unknown objects, which does not require prior knowledge. CFP-SLAM \cite{hu2022cfp} detects moving objects and uses a Kalman filter with a Hungarian algorithm to compensate for object misdetections. These methods obtain high accuracy in dynamic environments, and their results are considered state-of-the-art on the TUM RGB-D benchmark \cite{sturm2012benchmark}. While the DefSLAM \cite{lamarca_defslam_2021} tracks deformable objects with a parametric template, which limits the method to a small number of objects that are matched with the template, we are seeking a non-parametric method to detect deformable objects, paving the way to detect the intra-moving objects.
Nevertheless, their flip side is to represent moving objects as center points of segmentation blobs or bounding boxes, which are easily distorted by different views. Another scenario in which these methods fail is when only a part of the object moves. Still, its center remains unchanged (\textit{i.e.}, a person wobbles his head without displacing him from one place to another or rotating a revolving chair). These motions within an object cause significant noise to the feature extractor, leading to feature inconsistency between frames and, eventually, errors in camera pose estimation.

To address the above issues, we propose \textit{V3D-SLAM}, which differentiates moving objects from static objects by projecting depths into point clouds and analyzing their 3D shapes and geometry. \textit{V3D-SLAM} first identifies potentially moving objects by a novel Hough voting mechanism via the topology of 3D objects in one frame, resulting in two set objects (static objects and moving objects), thus analyzing the moving parts within static objects to seek moving parts via measuring the similarity using Chamfer distances \cite{butt1998optimum}.

\section{Related Work}

\textbf{RGB-D SLAM in Dynamic Environments}: Estimating poses of a moving camera has been an interesting topic in robotic vision, with the RGB-based earliest example using Tomasi–Kanade factorization \cite{tomasi1992shape}. This has laid the foundations for later research addressing dynamic objects in SLAM, including associating correspondences between RGB images \cite{kerl2013dense, mur2017orb, scona2018staticfusion}, weighting edge-like features for tracking \cite{li2017rgb}, identifying objects' movements via differences of consecutive frames \cite{sun2017improving}, differentiating static and dynamic objects via feature correlations \cite{kim2016effective, dai2020rgb}. However, these methods only deal with feature extraction on consecutive frames but lack prior semantic knowledge of the scene. To enhance this, recent works in vSLAM leverage object detection \cite{zhong2018detect, xiao2019dynamic, hu2022cfp} and segmentation \cite{yu2018ds, bescos2018dynaslam, yuan2020sad, ji2021towards, fan2022blitz, wang2022drg} models to extract RGB-based knowledge of the scene and use depth frames to reconstruct the scene's structure. Indeed, these methods only concentrate on 2D object movement via epipolar geometry while not guaranteeing 3D intra-object movement, which may be visually unchanged on RGB images. Furthermore, DefSLAM \cite{lamarca_defslam_2021} uses the precomputed template to detect deformable objects \cite{khangnguyen2023}, leading to a lack of generality to other categories. To address this issue, we employ the non-parametric method, which re-associates depth frames to information extracted from colored frames and computes the similarity between the current and subsequent frames.


\textbf{Spatial-Reasoned Hough Voting for Dynamic Objects}: 
The voting mechanism in Hough transform \cite{hough1959machine}, searching for the parameter that casts the most votes where each sample votes for another in parameter spaces, has been used widely for object detection \cite{gall2013class} with explicit and implicit features. DEHV \cite{sun2010depth} provides the Hough-based probabilistic approach, in which each object class and location cast the votes from depth scales. Max-margin Hough transform \cite{maji2009object} is also introduced to indicate the important weights for possible local locations of the object center. PoseCNN \cite{xiang2017posecnn} uses Hough voting to predict object centers via network-based extracted features. Similarly, VoteNet \cite{qi2019deep} samples point clouds to extract point features where votes are gained from the point feature domain. In this work, we devise a similar approach where the 3D dynamic objects are voted through the topology with other presented 3D objects. Instead of learning implicitly by using a neural network, each object will cast votes directly from pure geometry information.

\section{Overview of \textit{V3D-SLAM}}
\label{sec:system_description}

\begin{figure*}[t]
    \centering
    \vspace{2pt}
    \includegraphics[width=0.80\linewidth]{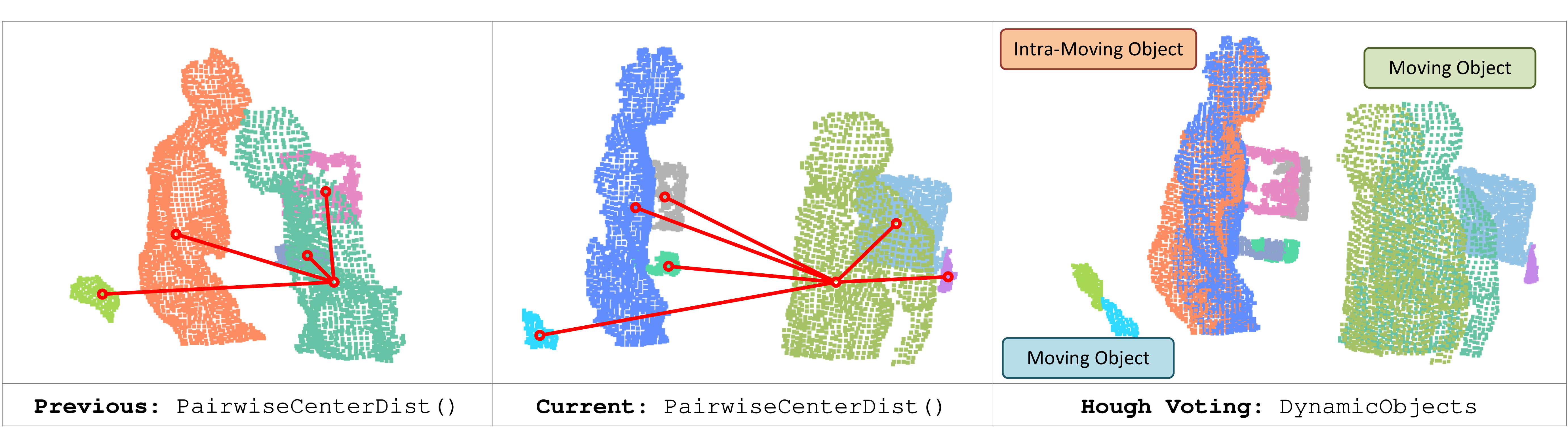}
    \vspace{-2pt}
    \caption{Spatial-reasoned Hough voting mechanism for moving objects (right) after computing accumulator array in previous (left) and current (middle) frames. Red lines illustrate Euclidean distances between one of the \texttt{`person'} objects and other presented entities. The moving \texttt{`person'} object and the \texttt{`chair'} object are identified using Alg. \ref{alg:voting_dynamic}; meanwhile, the other \texttt{`person'} object is only identified as an intra-moving object as he wobbles his head and his 3D centroid does not exceed the distance threshold for deformable objects.}
    \vspace{-18pt}
    \label{fig:voting_mechanism}
\end{figure*}

The overview of \textit{V3D-SLAM}, improving the robustness of dynamic RGB-D SLAM, is illustrated in Fig. \ref{fig:overview_method}. In between two consecutive frames, we first mask out potential moving objects from the scene \cite{yolov8, yolov9} to obtain static objects with backgrounds, followed by reconstructing point clouds of segmented instances. Noises induced by the segmentation model's uncertainty create fragments in point clouds of instance (Fig. \ref{fig:outlier_rejection}), which spatially do not belong to the objects, slightly shift the center of the point clouds, and influence the votes for potential dynamic objects significantly. To avoid these ambiguities, we statistically remove outliers at the instance level to eliminate segmentation-induced artifacts (Sec. \ref{sec:outlier_rejection}) before identifying instances' spatial centers. Thus, based on the topology of 3D objects between two frames, the voting mechanism (Fig. \ref{fig:voting_mechanism}) is implemented to identify moving objects with geometric and spatial information from themselves for other presented entities (Sec. \ref{sec:geometry_voting}).

In many indoor dynamic environments, the centers of large objects do not change from frame to frame. Still, some parts may be displaced, introducing another type of noise in the sequence. To identify these dynamic intra-objects, we track and calculate 3D objects' similarities via object deformation instead of objects' relative displacement on image planes due to their errors and distortion in different views (Sec. \ref{sec:object_deformation}). Within this tracking procedure, based on the physical constraints of objects' movements, we assume that the objects keep moving in the same direction in an infinitesimal period (between two frames in a 30-fps sequence) after identifying each moving object to compensate for the object misdetection or object that is out of field-of-view. Pixel-level features are also extracted on unmasked regions using ORB \cite{rublee2011orb} for camera pose estimation followed by camera trajectory optimization using pose graph optimization (PGO) \cite{teller2006fast} (Sec. \ref{sec:segmentation_extraction}). 

\section{Methodology}

\subsection{Instance Segmentation \& Feature Extraction}
\label{sec:segmentation_extraction}

We first segment the objects in RGB images using YOLOv8 \cite{yolov8}, then map the masked regions into depth images to make point clouds of instances. Hypothesizing as potentially moving, the objects, which are classified as \texttt{`person'}, are temporarily excluded from the scene to be re-evaluated for their movements in 3D space (Sec. \ref{sec:geometry_voting}). Thus, we detect edge- and corner-like features using ORB \cite{rublee2011orb} on unmasked regions containing static objects and the scene background. Not only for its segmenting capability, the segmentation model also gives us the number of objects, which benefits our subsequent noise removal procedure.

\subsection{Sensor Noises \& Segmentation Outlier Rejection} 
\label{sec:outlier_rejection}

In the context of SLAM, it is implicitly assumed that either the object or the robot is still. However, in dynamic environments, the fast-moving robot with the RGB-D camera produces perception uncertainty, which is best seen via RGB images, where the pixels are blurred, and depth maps, where the depth pixel cannot be interpolated from the previous frame \cite{scharstein2002taxonomy}. Furthermore, segmentation models add uncertainty when recognizing objects with RGB-D perception. To alleviate this, we categorize errors into two types: (1) depth map noises and (2) segmentation-induced mapping errors.

\subsubsection{Depth Map Noises} We re-estimate the depth value, $d(i, j)$, on depth maps via a $k \times k$ Gaussian kernel, $G(i, j)$, centered at that pixel to prevent potential errors caused by captured devices when calculating objects' center locations. The 2D-3D re-projection discretization is re-fined as below:
\begin{align*}
    z &= \int_{i=u-k}^{u+k} \int_{j=v-k}^{v+k}d(i, j) \cdot G(u, v)\\ &= \sum_{i=u-k}^{u+k} \sum_{j=v-k}^{v+k}d(i, j) \cdot \frac{1}{2\pi \sigma^2}e^{-{(u^2 + v^2)}/{2\sigma^2}} \\[0.125cm]
    x &= \frac{u - c_x}{f_x}\cdot z \text{ and } y = \frac{v - c_y}{f_y}\cdot z
\end{align*}
where $(u, v)$ represents the image coordinates of pixels, $(x, y, z)$ are their 3D coordinates, $(c_x, c_y, f_x, f_y)$ are defined as the camera's optical center and focal lengths, and $\sigma^{2}$ is the variance of $k^2$ pixels locating in the $k \times k$ Gaussian kernel.

\subsubsection{Segmentation-Induced Mapping Errrors} 
\begin{figure}[t]
    \centering
    \includegraphics[width=0.93\linewidth]{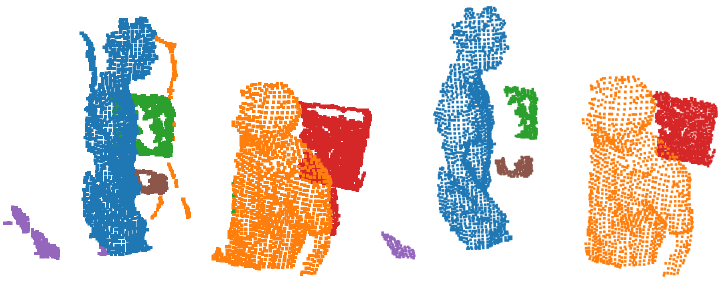}
    \vspace{-5pt}
    \caption{Outlier removal with semantic perception on point clouds.}
    \label{fig:outlier_rejection}
\end{figure}

As shown in Fig. \ref{fig:seg_feature}, RGB-based segmentation introduces pixel-level sensitivity when projecting image blobs via the corresponding depth maps to create point clouds of instances. Therefore, we first downsample point clouds through voxelization to efficiently remove these artifacts to guarantee balance and fairness during noise removal between dense and spare regions, as shown in Fig. \ref{fig:outlier_rejection} (right). Thus, we use density-based spatial clustering of applications with noises (DBSCAN) algorithm \cite{ester1996density} to group points into different clusters and hence filter out clusters by taking the $n$-first 3D large blobs with $n$ is the number of objects recognized provided by the segmentation model in Sec. \ref{sec:segmentation_extraction}, resulting in a finer point clouds of 3D semantic objects, as depicted in Fig. \ref{fig:outlier_rejection} (right).

\subsection{Spatial-Reasoned Hough Voting for Dynamic Objects}
\label{sec:geometry_voting}
Identifying dynamic objects in 2D images is complicated and sensitive to the dynamics of the environment, especially when the camera and objects move together since the object-to-object displacements are not correctly spatially interpreted with different projections from a moving camera. For example, objects can move at the same velocity as cameras, resulting in unchanged dynamic interpretations based solely on RGB images.

\vspace{-8pt}
\setlength{\textfloatsep}{0pt}
\begin{algorithm}[h]
    \caption{Hough Voting for Dynamic Objects}
    \label{alg:voting_dynamic}
    \begin{small}
        \DontPrintSemicolon
        \SetKwInOut{KwIn}{Input}
        \SetKwInOut{KwOut}{Output}
        \SetKwFunction{FMain}{VoteDynamicObjects}
        \SetKwProg{Pn}{function}{}{}
        \KwIn{$\textbf{O}_{p} \coloneqq$ objects in previous frame\\
        $\textbf{O}_{c} \coloneqq$ objects in current frame\\
        $\mathcal{T}_{d}, \mathcal{T}_{v} \coloneqq$ distance and voting counts thresholds}
        \KwOut{$\textit{DynamicObjects} \coloneqq$ lists of voted dynamic objects}
        \Pn{\FMain{\textbf{O}$_{p}$, \textbf{O}$_{c}$, $\mathcal{T}_d$, $\mathcal{T}_v$}}{ 
            \textbf{keys}$_p$, \textbf{dist}$_p$ = \texttt{PairwiseCenterDist}(\textbf{O}$_p$) \\
            \textbf{keys}$_c$, \textbf{dist}$_c$ = \texttt{PairwiseCenterDist}(\textbf{O}$_c$) \\
            accumulator = \{\} \\ 
            \For{$k_p \in \textbf{keys}_p$}{
                \For{$k_c \in \textbf{keys}_c$}{
                    \If{$k_p = k_c$ \textbf{and} $dist(d_p[k_p], d_c[k_c]) \geq \mathcal{T}_{d}$}{
                        id = \texttt{ExtractObjectId}($\textbf{key}_{c}$) \\
                        \texttt{accumulator}[id] += 1
                    }
                }
            }
            $\textit{DynamicObjects} =$ \texttt{accumulator}.\texttt{where}(id$\geq \mathcal{T}_{v}$)\\
            \KwRet{DynamicObjects}
        }
    \end{small}
\end{algorithm}
\vspace{-8pt}

To alleviate these concerns, we interpret scene dynamics by computing Euclidean distances between centroids of 3D instances in the current frame and estimating the displacement of objects in the next frame. If the displacement is larger than the pre-defined threshold, the accumulator $V(\textbf{O}_{i} | \textbf{O}_{j}, r)$ casts the vote from object $j$ to object $i$ for the high frame rate, $r \approx 30$; on the other hand, low-fps sequences neither can track and recognize objects.
\begin{equation}
    \vspace{-5pt}
     V(\textbf{O}_{i} | \textbf{O}_{j}, r) = \sum_{j} \textbf{I}\left[\texttt{dist}(\textbf{O}_i,\textbf{O}_j|r)\right]
     \label{eq:voting_accumulator}
     \vspace{-2pt}
\end{equation}
where $\texttt{dist}(\cdot)$ represents the Euclidean distance between 3D objects' centroids and $\textbf{I}(\cdot)$ denotes the indicator function and is as follows with the defined distance threshold, $\mathcal{T}_{\text{d}}$:
\begin{equation*}
    \vspace{-6pt}
    \textbf{I}\left[\texttt{dist}(\textbf{O}_i,\textbf{O}_j|r)\right] = \left\{
    \begin{array}{ll}
        1 \text{, if } \texttt{dist}(\textbf{O}_i,\textbf{O}_j|r) \geq \mathcal{T}_{\text{d}} \\
        0 \text{, otherwise.}
    \end{array} \right. 
    \label{eq:indicator}
\end{equation*}

\begin{table*}[!b]
    \vspace{-5pt}
    \centering
    \caption{Comparisons of ATE between RGB-D SLAM techniques.}
    \resizebox{18.0cm}{!}{
        \begin{tabular}{c | c c | c c | c c | c | c c | c c | c c | c c} 
            \hline
            \multirow{2}{*}{Sequences} & \multicolumn{2}{c}{\textbf{ORB-SLAM2} \cite{mur2017orb}} & \multicolumn{2}{c}{\textbf{DS-SLAM} \cite{yu2018ds}} & \multicolumn{2}{c}{\textbf{DynaSLAM} \cite{bescos2018dynaslam}} & \multicolumn{1}{c}{\textbf{TRS} \cite{ji2021towards}} & \multicolumn{2}{c}{\textbf{Blitz-SLAM} \cite{fan2022blitz}} & \multicolumn{2}{c}{\textbf{CFP-SLAM}$^-$ \cite{hu2022cfp}} & \multicolumn{2}{c}{\textbf{CFP-SLAM} \cite{hu2022cfp}} & \multicolumn{2}{c}{\textbf{V3D-SLAM (Ours)}} \\
            \cline{2-16}
            & RMSE & SD & RMSE & SD & RMSE & SD & RMSE & RMSE & SD & RMSE & SD & RMSE & SD & RMSE & SD \\ 
            \hline
            
            fr3/s/xyz & \boldssign{0.0092} & \boldssign{0.0047} & -- & -- & 0.0127 & 0.0060 & 0.0117 & 0.0148 & 0.0069 & 0.0129 & 0.0068 & \bolddagger{0.0090} & \boldasterisk{0.0042} & \boldasterisk{0.0087} & \bolddagger{0.0043} \\
            
            fr3/s/half & 0.0192 & 0.0110 & -- & -- & 0.0186 & 0.0086 & 0.0172 & 0.0160 & 0.0076 & \bolddagger{0.0159} & \boldssign{0.0072} & \boldasterisk{0.0147} & \bolddagger{0.0069} & \boldasterisk{0.0147} & \boldasterisk{0.0066} \\
            
            fr3/s/static & 0.0087 & 0.0042 & 0.0065 & 0.0033 & -- & -- & -- & -- & -- & \boldssign{0.0061} & \bolddagger{0.0029} & \boldasterisk{0.0053} & \boldasterisk{0.0027} & \bolddagger{0.0058} & \boldssign{0.0031} \\
            
            fr3/s/rpy & \bolddagger{0.0195} & \bolddagger{0.0124} & -- & -- & -- & -- & -- & -- & -- & \boldssign{0.0244} & \boldssign{0.0175} & 0.0253 & 0.0154 & \boldasterisk{0.0169} & \boldasterisk{0.0101} \\
            
            fr3/w/xyz & 0.7214 & 0.2560 & 0.0247 & 0.0161 & 0.0164 & 0.0086 & 0.0194 & \boldssign{0.0153} & 0.0078 & \bolddagger{0.0149} & \bolddagger{0.0077} & \boldasterisk{0.0141} & \boldasterisk{0.0072} & \boldssign{0.0153} & \boldssign{0.0080} \\
            
            fr3/w/half & 0.4667 & 0.2601 & 0.0303 & 0.0159 & 0.0296 & 0.0157 & 0.0290 & 0.0256 & 0.0126 & \bolddagger{0.0235} & \boldasterisk{0.0114} & \boldssign{0.0237} & \boldasterisk{0.0114} & \boldasterisk{0.0229} & \boldasterisk{0.0114} \\
            
            fr3/w/static & 0.3872 & 0.1636 & 0.0081 & 0.0036 & \boldssign{0.0068} & \boldssign{0.0032} & 0.0111 & 0.0102 & 0.0052 & 0.0069 & \boldssign{0.0032} & \bolddagger{0.0066} & \bolddagger{0.0030} & \boldasterisk{0.0065} & \boldasterisk{0.0028} \\
            
            fr3/w/rpy & 0.7842 & 0.4005 & 0.4442 & 0.2350 & \boldasterisk{0.0354} & \boldasterisk{0.0190} & \boldssign{0.0371} & \bolddagger{0.0356} & \bolddagger{0.0220} & 0.0411 & 0.0250 & 0.03680 & \boldssign{0.0230} & 0.0781 & 0.0360 \\
            
            \hline
        \end{tabular}
    }
    \label{tab:ate}
\end{table*}

Alg. \ref{alg:voting_dynamic} takes in objects, \textbf{O}, in previous (p) and current (c) frames as inputs and outputs the keys and IDs of objects that are voted as moving. Specifically, a class and object ID are assigned for each presented object instance in the frame. This information is tracked until the objects are no longer given in the scene. In contrast, the 3D pairwise distances of objects' centers are calculated among in-frame objects, and votes are cast among objects based on Eq. \ref{eq:voting_accumulator}.

\subsection{Intra-Object Movements}
\label{sec:object_deformation}
Objects like revolving chairs or people sitting without shifting their locations can cause noises by moving their heads or rotating the chair while talking with others. We treat these objects as deformable objects that must be considered when extracting their keypoints. To solve this, we use Chamfer distance to measure the similarity between two point clouds of segmented moving objects, $\mathcal{D}(\mathcal{P}_{1}, \mathcal{P}_{2})$, in consecutive frames:
\vspace{-5pt}
\begin{dmath}
    \mathcal{D}(\mathcal{P}_{1}, \mathcal{P}_{2}) = \frac{1}{|\mathcal{P}_1|}\left[\sum_{\textbf{p}_i \in \mathcal{P}_1} \minimize_{\textbf{p}_j \in \mathcal{P}_2} \left|\left|\textbf{p}_i - \textbf{p}_j\right|\right|_{2}^2\right] + \frac{1}{|\mathcal{P}_2|}\left[\sum_{\textbf{p}_j \in \mathcal{P}_2} \minimize_{\textbf{p}_i \in \mathcal{P}_1} \left|\left|\textbf{p}_j - \textbf{p}_i\right|\right|_{2}^2\right]
    \label{eq:similarity}
\end{dmath}
where $\mathcal{P}_1$ and $\mathcal{P}_2$ represent two clouds, respectively, $\textbf{p}_i$ and $\textbf{p}_j$ are $i^{th}$ and $j^{th}$ points in $\mathcal{P}_1$ and $\mathcal{P}_2$, respectively, and $\left|\mathcal{P}_{i}\right|$ indicates the number of points in the point cloud $\mathcal{P}_{i}$.

Using Eq. \ref{eq:similarity}, if an object is calculated as the deformable object as such definition, we also included features extracted from their RGB frame for camera pose estimation.

\subsection{Camera Pose Estimation}
After identifying dynamic objects, we mask them out of regions of interest and consider them as noises on images, leaving images constructed by static objects by changing the camera views. By extracting and tracking features on static objects and the scene background when the camera moves, we thus estimate the camera poses and trajectory. Inspired by the robustness of this task from ORB-SLAM2 \cite{mur2017orb} and ORB-SLAM3 \cite{campos2021orb}, we apply feature extraction on unmasked RBG images to extract keypoints on two consecutive frames, then matching the corresponding features between two keypoint sets. At this point, we assume two point sets are static point sets with different camera views. The transformation matrices are computed using RANSAC while removing outliers on these pointsets that do not satisfy the triangulation constraint of epipolar geometry of a pair RGBD. The two new point sets are used to look at depth to reconstruct 3D point sets, which are used to estimate the camera pose, including translation and rotation, using the RANSAC method. The estimated camera pose is added to the trajectory for closure detection and trajectory optimization using PGO \cite{teller2006fast}. 

\section{Evaluation on TUM RGB-D Bechmark}
\label{sec:evaluation}

\begin{figure*}[b]
    \vspace{-15pt}
    \centering
    \includegraphics[width=1.00\linewidth]{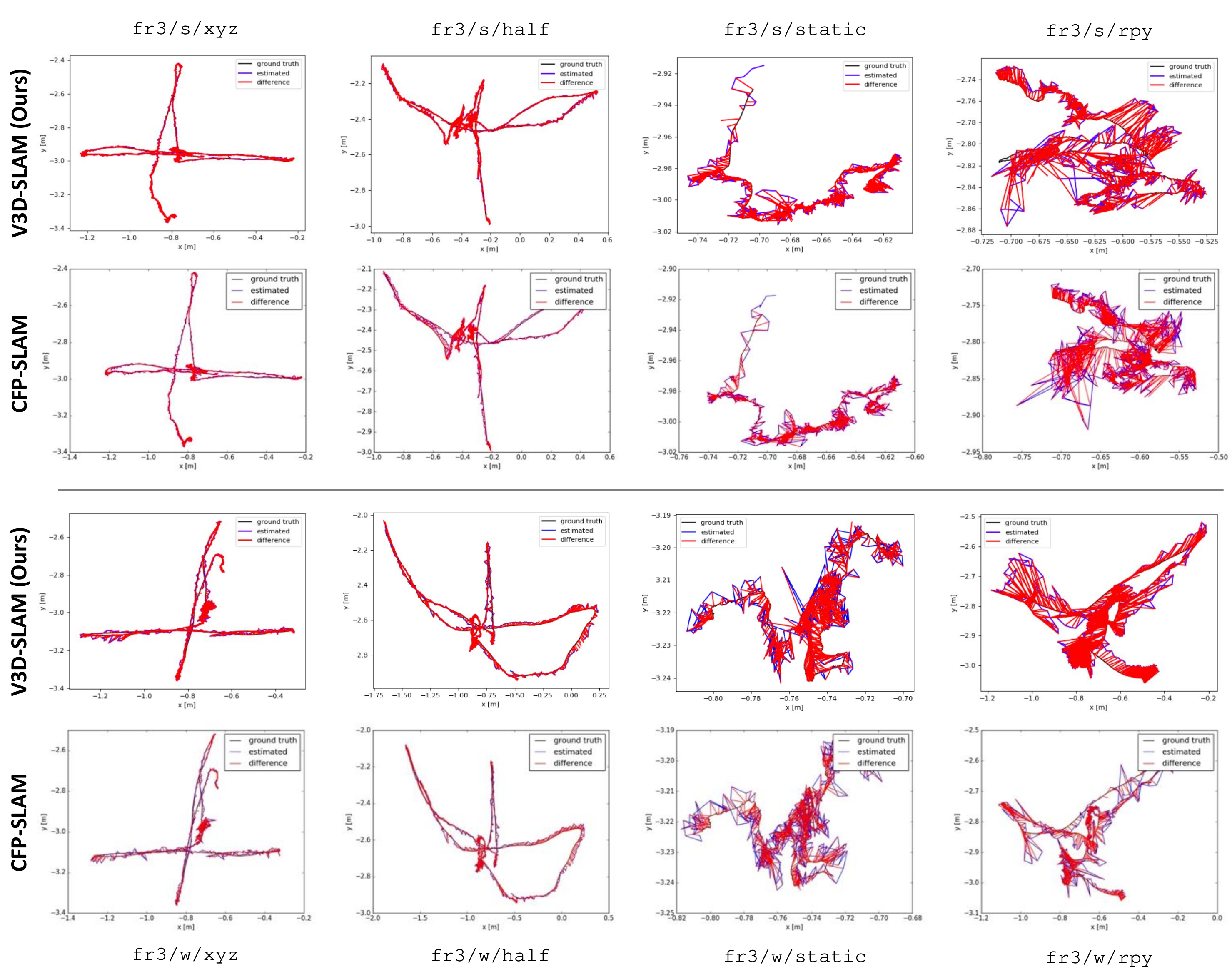}
    \vspace{-13pt}
    \caption{Qualitative results of camera trajectories of TUM RGB-D dynamic sequences estimated by our method and CFP-SLAM \cite{hu2022cfp}. The ground truth, the estimated trajectory, and their differences are encoded as black lines, blue lines, and red lines, respectively.}
    \vspace{-3pt}
    \label{fig:trajectory}
\end{figure*}

\begin{table*}[b!]
    \centering
    \resizebox{18.0cm}{!}{
        \begin{tabular}{c | c c | c c | c c | c | c c | c c | c c | c c} 
            \hline
            \multirow{2}{*}{Sequences} & \multicolumn{2}{c}{\textbf{ORB-SLAM2} \cite{mur2017orb}} & \multicolumn{2}{c}{\textbf{DS-SLAM} \cite{yu2018ds}} & \multicolumn{2}{c}{\textbf{DynaSLAM} \cite{bescos2018dynaslam}} & \multicolumn{1}{c}{\textbf{TRS} \cite{ji2021towards}} & \multicolumn{2}{c}{\textbf{Blitz-SLAM} \cite{fan2022blitz}} & \multicolumn{2}{c}{\textbf{CFP-SLAM}$^-$ \cite{hu2022cfp}} & \multicolumn{2}{c}{\textbf{CFP-SLAM} \cite{hu2022cfp}} & \multicolumn{2}{c}{\textbf{V3D-SLAM (Ours)}} \\
            \cline{2-16}
            & RMSE & SD & RMSE & SD & RMSE & SD & RMSE & RMSE & SD & RMSE & SD & RMSE & SD & RMSE & SD \\ 
            \hline
            
            fr3/s/xyz & \boldssign{0.0117} & \boldssign{0.0060} & -- & -- & 0.0142 & 0.0073 & 0.0166 & 0.0144 & 0.0071 &  0.0149 & 0.0081 & \bolddagger{0.0114} & \bolddagger{0.0055} & \boldasterisk{0.0105} & \boldasterisk{0.0051} \\
            
            fr3/s/half & 0.0231 & 0.0163 & -- & -- & 0.0239 & 0.0120 & 0.0259 & \bolddagger{0.0165} & \boldasterisk{0.0073} & 0.0214 & 0.0099 & \boldasterisk{0.0162} & \bolddagger{0.0079} & \boldssign{0.0184} & \boldssign{0.0088} \\
            
            fr3/s/static & 0.0090 & 0.0043 & \boldssign{0.0078} & 0.0038 & -- & -- & -- & -- & -- & \boldssign{0.0078} & \boldasterisk{0.0034} & \bolddagger{0.0072} & \bolddagger{0.0035} & \boldasterisk{0.0068} & \boldssign{0.0037} \\
            
            fr3/s/rpy & \bolddagger{0.0245} & \bolddagger{0.0144} & -- & -- & -- & -- & -- & -- & -- & 0.0322 & 0.0217 & \boldssign{0.0316} & \boldssign{0.0186} & \boldasterisk{0.0221} & \boldasterisk{0.0127} \\
            
            fr3/w/xyz & 0.3944 & 0.2964 & 0.0333 & 0.0229 & 0.0217 & 0.0119 & 0.0234 & 0.0197 & \boldasterisk{0.0096} & \boldssign{0.0196} & \boldssign{0.0099} & \boldasterisk{0.0190} & \bolddagger{0.0097} & \bolddagger{0.0193} & 0.0100 \\
            
            fr3/w/half & 0.3480 & 0.2859 & 0.0297 & 0.0152 & 0.0284 & 0.0149 & 0.0423 & \bolddagger{0.0253} & \bolddagger{0.0123} & 0.0274 & 0.0130 & \boldssign{0.0259} & \boldssign{0.0128} & \boldasterisk{0.0242} & \boldasterisk{0.0115} \\
            
            fr3/w/static & 0.2349 & 0.2151 & 0.0102 & 0.0048 & 0.0089 & 0.0044 &  0.0117 & 0.0129 & 0.0069 & \boldssign{0.0092} & \boldssign{0.0043} & \bolddagger{0.0089} & \bolddagger{0.0040} & \boldasterisk{0.0078} & \boldasterisk{0.0035} \\
            
            fr3/w/rpy & 0.4582 & 0.3447 & 0.1503 & 0.1168 & \boldasterisk{0.0448} & \boldasterisk{0.0262} & \bolddagger{0.0471} & 0.0473 & \bolddagger{0.0283} & 0.0540 & 0.0350 & \boldssign{0.0500} & \boldssign{0.0306} & 0.0726 & 0.0480 \\
            
            \hline
        \end{tabular}
    }
    \caption{Comparisons of Translational Drift in RPE between RGB-D SLAM techniques.}
    \label{tab:translational_rpe}
\end{table*}

\begin{table*}[b!]
    \centering
    \resizebox{18.0cm}{!}{
        \begin{tabular}{c | c c | c c | c c | c | c c | c c | c c | c c} 
            \hline
            \multirow{2}{*}{Sequences} & \multicolumn{2}{c}{\textbf{ORB-SLAM2} \cite{mur2017orb}} & \multicolumn{2}{c}{\textbf{DS-SLAM} \cite{yu2018ds}} & \multicolumn{2}{c}{\textbf{DynaSLAM} \cite{bescos2018dynaslam}} & \multicolumn{1}{c}{\textbf{TRS} \cite{ji2021towards}} & \multicolumn{2}{c}{\textbf{Blitz-SLAM} \cite{fan2022blitz}} & \multicolumn{2}{c}{\textbf{CFP-SLAM}$^-$ \cite{hu2022cfp}} & \multicolumn{2}{c}{\textbf{CFP-SLAM} \cite{hu2022cfp}} & \multicolumn{2}{c}{\textbf{V3D-SLAM (Ours)}} \\
            \cline{2-16}
            & RMSE & SD & RMSE & SD & RMSE & SD & RMSE & RMSE & SD & RMSE & SD & RMSE & SD & RMSE & SD \\ 
            \hline
            
            fr3/s/xyz & \boldssign{0.4890} & 0.2713 & -- & -- & 0.5042 & 0.2651 & 0.5968 & 0.5024 & \bolddagger{0.2634} & 0.5126 & 0.2793 & \bolddagger{0.4875} & \boldssign{0.2640} & \boldasterisk{0.4825} & \boldasterisk{0.2577} \\
            
            fr3/s/half & \boldssign{0.6015} & \boldssign{0.2924} & -- & -- & 0.7045 & 0.3488 & 0.7891 & \bolddagger{0.5981} & \boldasterisk{0.2739} & 0.7697 & 0.3718 & \boldasterisk{0.5917} & \bolddagger{0.2834} & 0.6543 & 0.3420 \\
            
            fr3/s/static & 0.2850 & 0.1241 & 0.2735 & 0.1215 & -- & -- & -- & -- & -- & \boldssign{0.2749} & \bolddagger{0.1192} & \boldasterisk{0.2654} & \boldasterisk{0.1183} & \bolddagger{0.2658} & \boldssign{0.1199} \\
            
            fr3/s/rpy & \boldssign{0.7772} & \boldssign{0.3999} & -- & -- & -- & -- & -- & -- & -- &  0.8303 & 0.4653 & \bolddagger{0.7410} & \bolddagger{0.3665} & \boldasterisk{0.6957} & \boldasterisk{0.3405} \\
            
            fr3/w/xyz & 7.7846 & 5.8335 & 0.8266 & 0.5826 & 0.6284 & 0.3848 & 0.6368 & \boldssign{0.6132} & \boldasterisk{0.3348} & 0.6204 & 0.3850 & \boldasterisk{0.6023} & \bolddagger{0.3719} & \bolddagger{0.6079} & \boldssign{0.3757} \\
            
            fr3/w/half & 7.2138 & 5.8299 & 0.8142 & 0.4101 & \boldssign{0.7842} & 0.4012 & 0.9650 & 0.7879 & \boldssign{0.3751} & 0.7853 & 0.3821 & \bolddagger{0.7575} & \bolddagger{0.3743} & \boldasterisk{0.6995} & \boldasterisk{0.3350} \\
            
            fr3/w/static & 4.1856 & 3.8077 & 0.2690 & 0.1182 & 0.2612 & 0.1259 & 0.2872 & 0.3038 & 0.1437 & \boldssign{0.2535} & \boldssign{0.1130} & \bolddagger{0.2527} & \bolddagger{0.1051} & \boldasterisk{0.2356} & \boldasterisk{0.1019} \\
            
            fr3/w/rpy & 8.8923 & 6.6658 & 3.0042 & 2.3065 & \boldasterisk{0.9894} & \bolddagger{0.5701} &  1.0587 & \boldssign{1.0841} & \boldssign{0.6668} & \bolddagger{1.0521} & \boldasterisk{0.5577} & 1.1084 & 0.6722 & 1.2773 & 0.8049 \\
            
            \hline
        \end{tabular}
    }
    \caption{Comparisons of Rotational Drift in RPE between RGB-D SLAM techniques.}
    \label{tab:rotational_rpe}
    \vspace{-10pt}
\end{table*}

\subsection{Testing Sequences \& Evaluation Metrics}
To test the performance of our proposed technique with spatial-reasoned votes for dynamic 3D objects, we evaluate Alg. \ref{alg:voting_dynamic} on the TUM RGB-D benchmark, containing eight dynamic sequences in terms of Absolute Trajectory Error (ATE) and Relative Pose Error (RPE), including translational and rotational drifts, in both root mean square error (RMSE) and standard deviation (SD). Across these metrics, we also compare our method against state-of-the-art vSLAM techniques: ORB-SLAM2 \cite{mur2017orb}, DS-SLAM \cite{yu2018ds}, DynaSLAM \cite{bescos2018dynaslam}, TRS \cite{ji2021towards}, Blitz-SLAM \cite{fan2022blitz}, and CFP-SLAM \cite{hu2022cfp}. 

\begin{figure*}[t]
    \centering
    \includegraphics[width=0.95\linewidth]{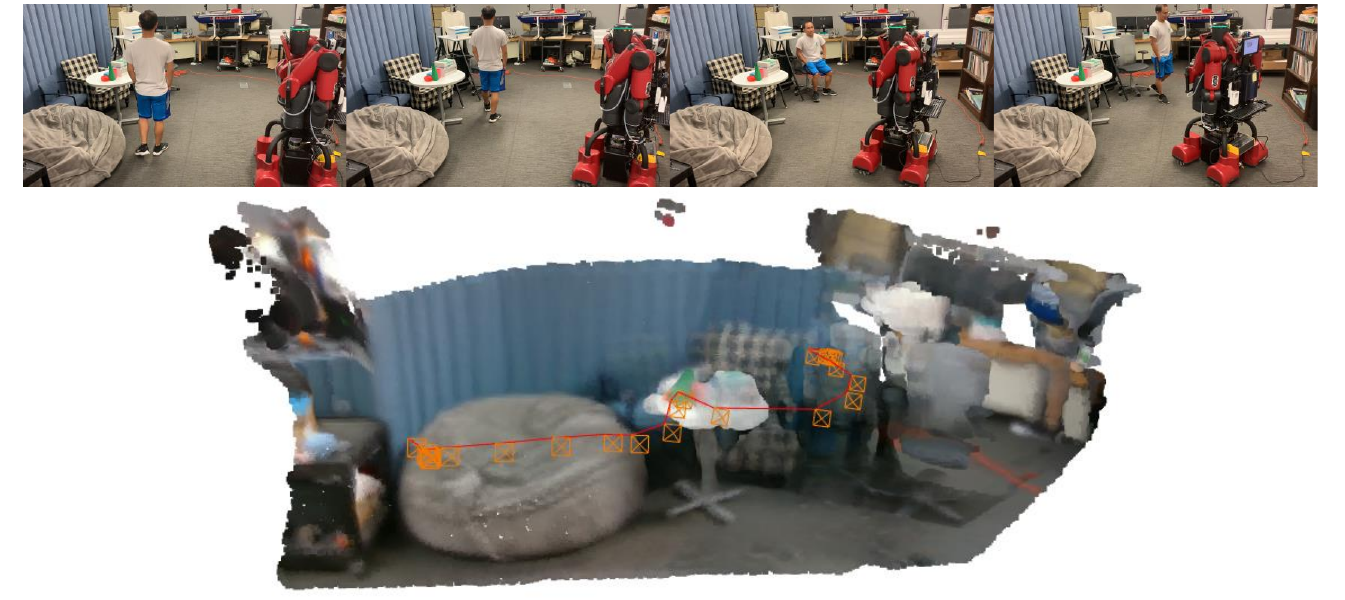}
    \vspace{-5pt}
    \caption{Moving sequences of the Baxter robot (top) and the resulting reconstructed 3D scene with the estimated camera trajectory (down). The red line indicates the estimated camera trajectory with optimization, and orange frustums mark camera poses along the trajectory.}
    \vspace{-18pt}
    \label{fig:experiments}
\end{figure*}

\subsection{Quantitative Results}
Table \ref{tab:ate} shows the quantitative comparisons between our method and other methods for the metric of ATE: for most of the sequences, we are able to obtain lower RMSE and SD compared to the recent state-of-the-art CFP-SLAM, except \texttt{fr3/w/xyz} and \texttt{fr3/w/rpy}. Table \ref{tab:translational_rpe} and Table \ref{tab:rotational_rpe} depict the RPE in terms of the camera's translational and rotational drifts, respectively. We achieve better estimates for translational drift, but the \texttt{fr3/s/half} sequence, where we underperform Blitz-SLAM and CFP-SLAM. For rotational drift, we obtain finer results in \texttt{fr3/s/xyz}, \texttt{fr3/s/rpy}, \texttt{fr3/w/half}, and \texttt{fr3/w/static} sequences, which are more significant compared to ORB-SLAM2 and CFP-SLAM.  For notations, we use  \boldasterisk{bold aterisk}, \bolddagger{bold dagger}, and \boldssign{bold section sign} to highlight the lowest error, the second-lowest error, and the third- error, respectively. 

\subsection{Qualitative Results}
Fig. \ref{fig:trajectory} shows the ATE along the camera trajectories for further assessment of qualitative results from our method (\textit{first and third lines}) and CFP-SLAM (\textit{second and fourth lines}). Fig. \ref{fig:trajectory} shows the ground truth (black), the estimated trajectory (blue), and their difference (red). Overall, our estimated trajectories have better ground-truth coverage than CFP-SLAM in most cases, with improvements in noticeable regions, but only worse for the \texttt{fr3/w/rpy} sequence.

\subsection{Ablation Study}
We use ORB-SLAM2 as a baseline where no moving objects and intra-objects are detected. The baseline code extracts features from images and generates spatial coordinates for mapping while adding keyframes and performing bundle adjustments using Graph Pose Optimization. Without Hough Voting, which predicts and removes the moving objects, performance gains by $8\%$ compared to the baseline but drops four times compared to full \textit{V3D-SLAM.} Without intra-object moving, it does not consider deformable parts/objects but moving objects, resulting in a dropping performance of $6\%$ compared to \textit{V3D-SLAM}.


\section{Real-Robot Experiments \& Deployability}
\label{sec:experiments}

\subsection{Experimental Setup} 
The Intel RealSense D435i RGB-D camera is mounted on the display of the Baxter robot, and the scene of various indoor objects with bean bags, backpacks, chairs, books, and cups is set up, as shown in Fig. \ref{fig:experiments}. The robot perceives objects within a 3-meter radius, the ideal range given by the camera, while arbitrarily moving to estimate its camera poses and trajectory and reconstruct the scene simultaneously with the acquired RGB-D stream.

\subsection{Performance of Proposed Method}
We deploy our method on the Intel NUC5i3RYH PC, which runs on its native onboard CPU without any dedicated GPUs. The entire pipeline sufficiently achieves on-robot deployable performance. Fig. \ref{fig:experiments} shows the arbitrary movements of the Baxter robot to perceive the setup scene up front at the top row. Our arbitrary movements supported by four mecanum wheels on the integrated Dataspeed mobility base allow the robot to slide to the left and right side, move toward and backward, and rotate in a clockwise and anti-clockwise manner, closely mimicking the camera's movements in the TUM RGB-D benchmark \cite{sturm2012benchmark}. The qualitative results of the estimated camera trajectory in 3D and the 3D reconstructed scene are also shown in Fig. \ref{fig:experiments} at the bottom row with the red line indicating the estimated camera trajectory with optimization, and the camera frustums are marked in orange.

\vspace{-1pt}
\subsection{Demonstration}
The demonstration video shows the deployability of our method on the Baxter mobile robot with an employed Intel RealSense D435i RGB-D camera and is made available at \url{https://youtu.be/K4RcKrASpqI}.

\section{Conclusions}

In this paper, we present \textit{V3D-SLAM}, a technique that reliably estimates and reconstructs the camera trajectory by removing noises induced by the dynamic nature of the environment. The dynamic objects are identified at the pixel level using state-of-the-art object segmentation on RGB images and refined using geometrical information in the 3D domain. Besides using Hough voting to identify moving objects in 3D, we also detect deformable objects using Chamfer distance to exclude their intra-object changes, which noise the feature extractor. To verify the robustness of our proposed method, we conduct experiments on the TUM RGB-D benchmark and compare our proposed method to recent state-of-the-art SLAM techniques. The experimental results show that our method mostly outperforms others regarding ATE and rotational and translational RPE metrics. Through deployment, our method enables the Baxter robot to perform RGB-D SLAM and its operations.

\pagebreak
\bibliographystyle{IEEEtran}
\bibliography{IEEEabrv, 09_references}

\end{document}